\documentclass{interact}

\usepackage{natbib}
\bibpunct[, ]{(}{)}{;}{a}{}{,}

\usepackage{multirow}
\usepackage[nolists,tablesfirst]{endfloat}
\DeclareDelayedFloatFlavor{sidewaystable}{table}

\usepackage[misc]{ifsym}

\begin{document}

\articletype{Review}

\title{Robotic Process Automation - A Systematic Literature Review and Assessment Framework}

\author{
\name{Judith Wewerka\textsuperscript{a \Letter} and Manfred Reichert\textsuperscript{a}}
\affil{\textsuperscript{a} Institute of Databases and Information Systems, Ulm University, Germany}
\email{\{judith.wewerka, manfred.reichert\}@uni-ulm.de} }

\maketitle

\begin{abstract}
Robotic Process Automation (RPA) is the automation of rule-based routine processes to increase efficiency and to reduce costs. Due to the utmost importance of process automation in industry, RPA attracts increasing attention in the scientific field as well. This paper presents the state-of-the-art in the RPA field by means of a Systematic Literature Review (SLR). In this SLR, 63 publications are identified, categorised, and analysed along well-defined research questions. From the SLR findings, moreover, a framework for systematically analysing, assessing, and comparing existing as well as upcoming RPA works is derived. The discovered thematic clusters advise further investigations in order to develop an even more detailed structural research approach for RPA.
\end{abstract}


\section{Introduction}
In our continuously changing world, it is indispensable that business processes are highly adaptive \citep{Reichert2012} and become more efficient and cost-effective \citep{Lohrmann2016}. As a consequence, companies demand for an increasing degree of process automation to stay competitive in their markets. In this context, the use of software robots (bots for short) mimicking human interaction, also denoted as Robotic Process Automation (RPA), constitutes a `highly promising approach' \citep{Cewe2017} and more and more companies rely on this cutting edge technology \citep{Asatiani2016} to optimise and implement their internal business processes.

\subsection{Problem Statement}
\label{ch:ps}
RPA constitutes an emerging technology raising high expectations in industry \citep{Auth2019}. For companies, however, it is still difficult to grasp the fundamental concepts of RPA, to understand the differences in comparison to other methods and technologies (e.g., Business Process Management, BPM), and to estimate the effects the introduction of RPA will have on the company and its employees.

Due to the increasing scientific attention of RPA, the number of publications on RPA will further increase over time. This necessitates a framework for analysing, assessing, and comparing these works. Therefore, the systematic literature review (SLR) presents relevant RPA publications and the results are used to derive such a framework.

\subsection{Contribution}
This paper provides an SLR on RPA as well as an assessment framework derived from it. First, we aim to present the state-of-the-art of RPA by systematically analysing and assessing the most relevant publications in the field. In this context, we provide RPA definitions, discuss differences to related technologies, introduce criteria for RPA-suitable business processes, and give insights into RPA effects. Further, we present case studies, give an overview of RPA methods, and discuss the combination of Artificial Intelligence (AI) with RPA. 
Second, taking the results of the SLR, we derive the ANCOPUR framework for systematically \textbf{an}alysing and \textbf{co}mparing emerging \textbf{pu}blications on \textbf{R}PA and link them to existing publications to finally assess their novelty and contribution to research.

The remainder of this paper is structured as follows. Section~\ref{ch:m} introduces the methodology, followed by the obtained results in Section~\ref{ch:results}. Section~\ref{ch:ANCOPUR} derives the ANCOPUR framework. Related work is presented in Section~\ref{ch:relatedwork}. Then, the results are discussed in Section~\ref{ch:discussion}. We conclude with a summary and an outlook in Section~\ref{ch:summary}.

\section{Methodology}
\label{ch:m}
An SLR is conducted to analyse both the body of knowledge and relevant publications in the RPA field. An SLR is `a means of identifying, evaluating and interpreting all available research relevant to a particular research question (RQ), or topic area, or phenomenon of interest' \citep{Kitchenham2004}. Following the guidelines described by Kitchenham, we design a protocol that describes the formulation of both research questions (cf. Section~\ref{ch:rq}) and the search string (cf. Section~\ref{ch:ss}), the identification of data sources (cf. Section~\ref{ch:ds}), the definition of inclusion and exclusion criteria (cf. Section~\ref{ch:inex}), the elaboration of quality assessment questions (cf. Section~\ref{ch:qa}), the selection of publications (cf. Section~\ref{ch:sepub}), the data extraction method (cf. Section~\ref{ch:daex}), and the data analysis method (cf. Section~\ref{ch:da}).

\subsection{Formulation of the Research Questions}
\label{ch:rq}

Our general goal is to analyse the body of relevant publications in the RPA field. In a first step, we want to understand the technology perspective of RPA and how it differs from related technologies, like intelligent or cognitive automation \citep{Bruno2017, Schmitz2019a, Suri2018} or BPM \citep{Cewe2017}. This results in our first research question:
\textbf{RQ 1: What is RPA and what are the differences between RPA and related technologies?}
Secondly, criteria for assessing whether or not a given business process or parts of it are suited for RPA are investigated. Furthermore, we are interested in the tools available for implementing RPA. This leads to our second research question:
\textbf{RQ 2: Which business processes can be automated with RPA and which tools are used for automation?}
For newly emerging technologies, like RPA, the question arises whether it is worthwhile to adapt it. Therefore, in a third step, we want to systematically understand RPA effects on humans and their work life as well as on the companies implementing RPA projects. This results in our third research question:
\textbf{RQ 3: What are RPA effects?}
In a fourth step, we investigate how far research has taken up on RPA. Particularly, we are interested in methods that aim to foster RPA implementation. This leads to our fourth research question:
\textbf{RQ 4: Are there methods for improving the implementation of RPA projects?}
Finally, the growing importance of AI in many areas raises the question to what degree AI plays a role in connection with intelligent process automation. The fifth research question addresses the topic of combining AI with RPA:
\textbf{RQ 5: Is AI used in combination with RPA?}

\subsection{Formulation of the Search String}
\label{ch:ss}

We elaborate the search string iteratively based on our knowledge of the topic, the pre-specified research questions, and pilot searches. The search string is refined to retrieve a maximum number of different publications. The pilot searches are inspected to ensure that all relevant publications are found. The final search string for the SLR is as follows:

\begin{center}
\textbf{`robotic process automation' OR `intelligent process automation' OR
`tools process automation' OR `artificial intelligence in business process' OR
`machine learning in business process' OR `cognitive process automation'.}
\end{center}

Note that the abbreviation `RPA' is not included, as the search then would yield around 31.000 results. RPA not only serves as acronym for Robotic Process Automation, but also for \textit{Recombinase Polymerase Amplification} in the field of DNA chemistry and others. Though we omit the acronym RPA, all relevant publications are still included in the results.

\subsection{Identification of Data Sources}
\label{ch:ds}

We apply the search string to different data sources to find relevant publications. Five electronic libraries are identified as relevant for conducting the SLR as they cover scientific publications in Computer Science:
\begin{center}
\textbf{ACM Digital Library, Science Direct - Elsevier, IEEE \\ Xplore Digital Library, SpringerLink, and Google Scholar.}
\end{center}
Additionally, we consider literature cited by the retrieved publications by performing a \textit{backward reference search} \citep{JalaliSamirehandWohlin2012}. Finally, Google Scholar alerts are analysed during the SLR procedure and the writing process to get notified about newly emerging publications on the topic.

\subsection{Definition of Inclusion and Exclusion Criteria}
\label{ch:inex}

To identify relevant publications, we define the following inclusion and exclusion criteria.

\textbf{Inclusion Criteria:}
\begin{itemize}
	\item[1.)] The publication deals with the topic of RPA and contributes answers to at least one of the research questions.
	\item[2.)] The title and the abstract seem to contribute to our research questions and contain terms such as robotic/intelligent/cognitive process automation, virtual assistant, process intelligence, business process model automation, intelligent business process management, or software bot.
\end{itemize}

\textbf{Exclusion Criteria:}
\begin{itemize}
	\item[1.)] The publication is not written in English.
	\item[2.)] The title and abstract do not seem to contribute to our research questions and contain words such as business process management, business intelligence, analytics, multi-agent system, big data, or process mining. 
	\item[3.)] The publication is a patent, master thesis, or web page.	
	\item[4.)] The publication is not electronically accessible without payment. 
	\item[5.)] All relevant aspects of the publication are included in another publication.
	\item[6.)] The publication only compares existing research and has no new input. 
\end{itemize}

A publication is included if both inclusion criteria are met, and it is excluded if any of the exclusion criteria is fulfilled.

\subsection{Elaboration of Quality Assessment Questions}
\label{ch:qa}

RPA is a relatively new research area (cf. Figure~\ref{img:dist} in Section~\ref{ch:results}). The topic is mostly driven by industry. Thus, applying rigid quality assessment questions would probably exclude relevant publications. Therefore, we decide against the introduction of additional quality criteria.

\subsection{Selection of Publications}
\label{ch:sepub}

The search string (cf. Section~\ref{ch:ss}) is applied to the identified data sources (cf. Section~\ref{ch:ds}), which yields 1510 results (Inclusion Criterion 1). To select relevant publications, the metadata is loaded into Microsoft Excel. It includes title, author, year, abstract, and keywords. In a first step, duplicates and publications not written in English (Exclusion Criterion 1) are excluded resulting in 1045 publications. Then, publications whose title does not indicate any contribution to one of the research questions are excluded, leaving 289 publications (Inclusion Criterion 2, Exclusion Criterion 2). Following this, the abstracts of the remaining publications are scanned leading to 201 publications (Inclusion Criterion 2, Exclusion Criterion 2). We then exclude publications corresponding to patents, theses, or web pages, resulting in 142 relevant publications (Exclusion Criterion 3). Thereof, 125 are accessible without payment (Exclusion Criterion 4) and 85 are not included in another publication (Exclusion Criterion 5). Finally, 39 publications provide new input to the research questions and are included in the final publication list (Exclusion Criterion 6). Through backward referencing one additional publication is identified and included. 

The initial search was performed on June 6th 2019. Since then (until June 2020) the alerts from Google Scholar have revealed 1206 new publications. 23 of them meet the inclusion criteria, but do not fulfil any of the exclusion criteria. Thus, they are added to our final publication list leading to 63 relevant publications.

\subsection{Data Extraction Method}
\label{ch:daex}

To each of the 63 relevant publications, a data extraction process is applied in order to answer the research questions derived in Section~\ref{ch:rq}. We extract the following information:

\begin{itemize}
	\item[1.)] General information, i.e., title, author, publication year, publication venue, number of citations, and publication type,
	\item[2.)] Definitions provided for RPA (RQ 1),
	\item[3.)] Differences between RPA and related technologies, e.g., intelligent automation, BPM, etc. (RQ 1),
	\item[4.)] Criteria for selecting suitable business processes for RPA (RQ 2),
	\item[5.)] Concrete business processes automated in specific business areas with an explicitly mentioned automation tool (RQ 2),
	\item[6.)] RPA effects on humans, work life, and companies (RQ 3),
	\item[7.)] Methods to improve RPA projects (RQ 4),
	\item[8.)] Combination of RPA with AI (RQ 5), and
	\item[9.)] Significant information outside the scope of the derived research questions.
\end{itemize}

Table~\ref{tab:pub} and its continuation Table~\ref{tab:pub2} give an overview of the 63 relevant publications indicating the reference, ID, title, type of publication, and research questions the publication refers to. In the following, the ID is used to reference the corresponding publication. The publication type distinguishes between \textit{Method}, \textit{Case Study}, \textit{Review}, and \textit{Research Paper}. A publication is classified as \textit{Method} if it reports on the development and testing of a new RPA method, as \textit{Case Study} if it focuses on a practical use case, as \textit{Review} if it provides a synthesis of acquainted knowledge, and as \textit{Research} otherwise. 

\begin{sidewaystable}
\setcounter{table}{0}
\tbl{Final list of the 63 relevant publications indicating reference, ID, and answers to the research questions.}
{\begin{tabular}{@{}p{5cm}lp{10cm}lccccc} \toprule
Ref & ID & Title & Type & RQ1 & RQ2 & RQ3 & RQ4 & RQ5 \\ \midrule
\citep{Aguirre2017} & P01 &Automation of a Business Process Using Robotic Process Automation (RPA): A Case Study & Case Study & x & x & x &  & \\ 
\citep{Asatiani2016} & P02 & Turning robotic process automation into commercial success - Case OpusCapita & Research & x & x & x & & \\
\citep{Asquith2019} & P03 &Let the robots do it! - Taking a look at Robotic Process Automation and its potential application in digital forensics & Case Study & x & x & x & & \\
\citep{Auth2019} & P04 &Impact of Robotic Process Automation on Enterprise Architectures &Research & & & x & & \\
\citep{Bosco2019} & P05 &Discovering Automatable Routines From User Interaction Logs & Method & & & & x & \\
\citep{Bruno2017} & P06 &Robotic disruption and the new revenue cycle & Research & x & x & x & & \\
\citep{Cewe2017} & P07 & Minimal Effort Requirements Engineering for Robotic Process Automation with Test Driven Development and Screen Recording & Method & x & & & x & \\
\citep{Chacon-Montero2019} & P08 & Towards a Method for Automated Testing in Robotic Process Automation Projects & Method & & & & x &  \\
\citep{Chalmers2019} & P09 & Machine Learning with Certainty: A Requirement For Intelligent Process Automation & Research & & & & & x\\
\citep{Cohen2019} & P10 &Exploring the Use of Robotic Process Automation (RPA) in Substantive Audit Procedures & Case Study & x & x & x & & \\
\citep{Eikebrokk2019} & P11 & Robotic Process Automation for Knowledge Workers - Will It Lead To Empowerment or Lay-Offs? & Research & & & x & & \\
\citep{Eikebrokk2020} & P12 & Robotic Process Automation and Consequences for Knowledge Workers; a Mixed-Method Study & Research & & & x & & \\
\citep{Fernandez2018} & P13 &Impacts of Robotic Process Automation on Global Accounting Services & Research & & & x & & \\
\citep{Fung2013} & P14 &Criteria, Use Cases and Effects of Information Technology Process Automation (ITPA) & Research & x & x & x & & \\
\citep{Gao2019} & P15 & Automated robotic process automation: A self-learning approach & Method & & & & x& \\
\citep{Geyer-Klingeberg2018} & P16 &Process Mining and Robotic Process Automation: A Perfect Match & Method & & & & x & \\
\citep{Hallikainen2018} & P17 &How OpusCapita Used Internal RPA Capabilities to Offer Services to Clients & Case Study & & x & x & & \\
\citep{Hindel2020} & P18 & Robotic Process Automation: Hype or Hope? & Research & & &x & & \\
\citep{Houy2019} & P19 &Robotic Process Automation in Public Administrations & Research & x & & x & & x \\
\citep{Huang2019} & P20 & Applying robotic process automation (RPA) in auditing: A framework & Method & & & & x & \\
\citep{Hwang2020} & P21 & MIORPA: Middleware System for open-source robotic process automation & Method & & & & x & \\
\citep{Ian2016} & P22 &The future of professional work: Will you be replaced or will you be sitting next to a robot? & Method & & & & x & \\
\citep{Issac2018} & P23 &Delineated Analysis of Robotic Process Automation Tools & Review & & x & & & \\ 
\citep{Jimenez-Ramirez2020} & P24 & Automated testing in robotic process automation projects & Method & & & & x & \\
\citep{Jimenez-Ramirez2019} & P25 &A Method to Improve the Early Stages of the Robotic Process Automation Lifecycle & Method & & & x & x & \\
\citep{Kin2018} & P26 &Cognitive Automation Robots (CAR) & Research & & & & & x \\
\citep{Koch2020} & P27 & `Mirror, Mirror, on the wall': Robotic Process Automation in the Public Sector using a Digital Twin & Method & & & & x & \\
\citep{Kokina2019} & P28 & Early evidence of digital labor in accounting: Innovation with Robotic Process Automation & Research & & x & x && \\
\citep{Lacity2017a} &P29 & A New Approach to Automating Services & Research & x & & x & & \\
\citep{Lacity2015a} & P30 &Robotic process automation at Xchanging & Case Study & & x & x & & \\
\citep{Lacity2015} & P31 &Robotic Process Automation: Mature Capabilities in the Energy Sector & Case Study & & x & x & & \\
\citep{Lacity2016} & P32 &Robotizing Global Financial Shared Services at Royal DSM & Case Study & & x & x & & \\
\citep{Lacity2017} & P33 &Service Automation: Cognitive Virtual Agents at SEB Bank & Research & x & & x & & x \\
\citep{Leno2020a} & P34 & Automated Discovery of Data Transformations for Robotic Process Automation & Method & & & & x & \\
\citep{Leno2018} &P35 &Multi-Perspective process model discovery for robotic process automation & Method & & & x & x & \\
\citep{Leno2020} & P36 & Robotic Process Mining: Vision and Challenges & Method & & & & x & \\
\citep{Leno2019} & P37 &Action Logger: Enabling Process Mining for Robotic Process Automation & Method & & & & x & \\
\citep{Leopold2018} & P38 &Identifying candidate tasks for robotic process automation in textual process descriptions & Method & & & & x & \\ \bottomrule
\end{tabular} }
\label{tab:pub}
\end{sidewaystable}

\begin{sidewaystable}
\setcounter{table}{1}
\tbl{Final list of the 63 relevant publications indicating reference, ID, and answers to the research questions - Continuation.}
{\begin{tabular}{@{}p{5cm}lp{10cm}lccccc} \toprule
Ref & ID & Title & Type & RQ1 & RQ2 & RQ3 & RQ4 & RQ5 \\ \midrule
\citep{Leshob2018} & P39 &Towards a Process Analysis Approach to Adopt Robotic Process Automation & Method & & & & x & \\
\citep{Lewicki2019} & P40 &Are Robots Taking Our Jobs? A RoboPlatform at a Bank & Case Study & x & x & & & x \\
\citep{Masood2019} & P41 &Cognitive Robotics Process Automation: Automate This! & Research & & & x & & x \\
\citep{Moffitt2018} & P42 &Robotic Process Automation for Auditing & Research & & & x & & \\
\citep{Mohanty2018} & P43 &Intelligent Process Automation = RPA + AI & Research & x & & x & & x \\
\citep{Osmundsen2019} & P44 &Organizing Robotic Process Automation: Balancing Loose and Tight Coupling & Method & x & & & x & \\
\citep{Patel2019} & P45 &Customized Automated Email Response Bot using Machine Learning and Robotic Process Automation & Research & x & & & & x \\
\citep{Penttinen2018} & P46 &How to Choose Between Robotic Process Automation and Back-End System Automation? & Research & x & x & x & & \\
\citep{Radke2020} & P47 & Using Robotic Process Automation (RPA) to enhance Item Master Data Maintenance Process & Case Study & & x & x & & \\
\citep{Riedl2019} & P48 &Robotic Process Automation: Developing a Multi-Criteria Evaluation Model for the Selection of Automatable Business Processes & Method & & & & x & \\
\citep{Rutschi2020} & P49 & Towards a framework of implementing software robots: Transforming Human-executed Routines into Machines & Method & & & & x & \\
\citep{Schmitz2019} & P50 &Enabling digital transformation through robotic process automation at Deutsche Telekom & Case Study & & x & x & & \\
\citep{Schmitz2019a} & P51 &Smart Automation as Enabler of Digitalization? A Review of RPA/AI Potential and Barriers to Its Realization & Review & x & x & x & & \\
\citep{Seguin2020} & P52 & Robotic Process Automation (RPA) Using an Integer Linear Programming Formulation & Method & & & & x & \\
\citep{Stople2017} &P53 & Lightweight IT and the IT Function: experiences from robotic process automation in a Norwegian bank& Case Study & x & x & & & \\
\citep{Suri2018} & P54 &Automation of Knowledge-Based Shared Services and Centers of Expertise & Research & x & & x & & \\
\citep{Suri2017} & P55 &Software Bots - The next frontier for shared services and functional excellence & Research & & & x & & \\
\citep{VanderAalst2018} &P56 & Robotic Process Automation & Research & x & & & & x \\
\citep{Wanner2020} & P57 & Process selection in RPA projects - Towards a quantifiable method of decision making & Method & & & & x & \\
\citep{Willcocks2016} & P58 &Robotic Process Automation: The Next Transformation Lever for Shared Services & Case Study & x & x & x & & \\
\citep{Willcocks2015a} &P59 & Robotic Process Automation at Telef{\'{o}}nica O2 & Case Study & x & x & x & & \\
\citep{Willcocks2015} & P60 &The IT Function and Robotic Process Automation & Research & x & & & & \\
\citep{William2019} & P61 & Improving Corporate Secretary Productivity Using Robotic Process Automation & Case Study & & x & x & & \\
\citep{Wroblewska2018} & P62 &Robotic Process Automation of Unstructured Data with Machine Learning & Research & & x & x & & x \\
\citep{Yatskiv2019} & P63 &Improved Method of Software Automation Testing Based on the Robotic Process Automation Technology & Case Study & & x & & & \\ \bottomrule
\end{tabular} } 
\label{tab:pub2}
\end{sidewaystable}

\subsection{Data Analysis Method}
\label{ch:da}

After having extracted relevant data from all selected publications, we cluster the obtained data. For each research question, we scan relevant information and build groups based on matches and differences. 

Concerning RQ 1, for example, we study all definitions provided by the publications, identify different aspects, e.g., `software-based solution', `mimics human behaviour' or `rule-based nature', and label the publications according to the aspects they cover. The same procedure is applied to bundle differences to other technologies (RQ 1), process selection criteria (RQ 2), and effects (RQ 3).

Depending on the publication type, different data analysis methods are then applied. For case studies, we investigate the business area, the concerned business process, and the used automation tool. Then, we cluster these case studies (RQ 2). Method papers are co-related with the stage of the RPA project, which they aim to improve, in order to identify common points (RQ 4). Finally, research papers answering RQ 5 are treated separately to group approaches for combining RPA with AI.

\section{Results}
\label{ch:results}

\begin{figure}
\centering
\includegraphics[width=\textwidth]{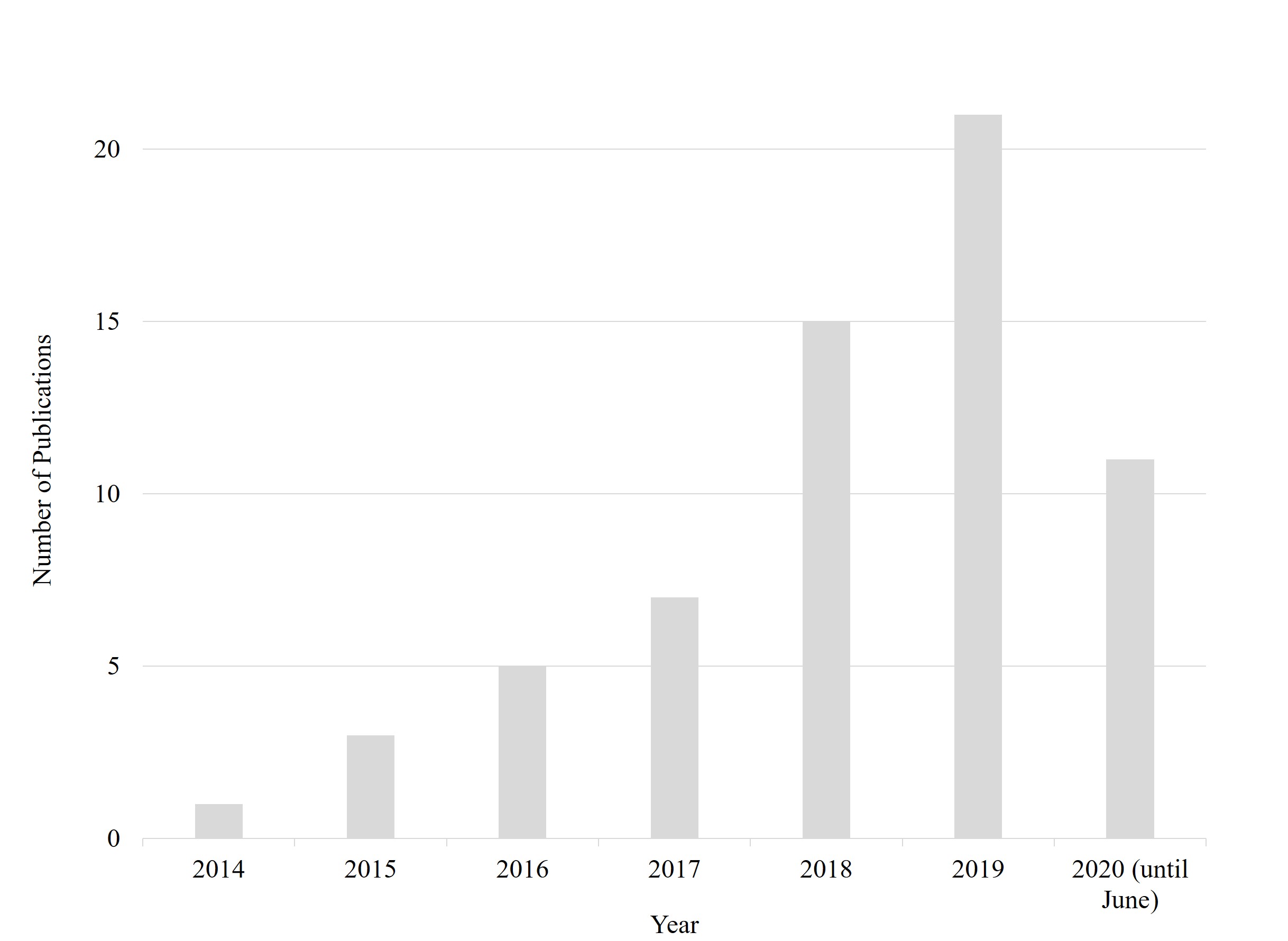}
\caption{Distribution of Publications over Years.}
\label{img:dist}
\end{figure}

In this chapter, we analyse the 63 publications identified by the SLR to answer the research questions described in Section~\ref{ch:rq}. The answers are structured along the research questions and the seven discovered thematic clusters.
In general, we have noticed a growing interest in RPA. Figure~\ref{img:dist} shows the distribution of the publications over the recent years; it started with one to seven publications in the years 2014 to 2017. In 2018, 15 relevant publications appeared and in 2019, 21 works were published. In 2020, until June, 11 publications could be identified.

Concerning the publication venue, there is no clear majority visible. RPA is important in a variety of areas covered by different conferences and journals. Regarding authorship, two researchers are dominating: M. Lacity and L. Willcocks are both (co-)authors of eight publications each.

The 63 publications comprise 15 case studies, 22 methods, two reviews, and 24 research papers.

\subsection{RQ 1: What is RPA and what are the differences between RPA and related technologies?}
\label{ch:rq1}

\textbf{Definition.} To better understand RPA, we look at the various definitions provided in literature. A first definition can be found in P60: `RPA is a software-based solution [...] [and] refers to configuring the software ``robot" to do the work previously done by people.' Already in 2014, P14 provided a definition, which referred to \textit{Information Technology Process Automation} instead of RPA. 

The definition from P60 addresses two aspects. First, RPA corresponds to a \textbf{software-based solution} (cf. P33, P51, P58, P59). Second, it \textbf{mimics human behaviour} (cf. P43, P45, P51, P53, P59). Most of the other definitions in literature pick up those aspects expanding it by mainly two other characteristics. Instead of the term `software-based solution', terms like `software robot' (P40, P43) or `virtual assistant' (P02) are used. Mimicking human behaviour is also expressed by phrases like `enters data, just as a human would' (P40), `mimic human actions' (P06), or `operate [...] in the way a human would do' (P56).

To augment the definition, characteristics of the automated processes are included. These characteristics cover their \textbf{rule-based nature} (P01, P10, P29, P45), the inclusion of \textbf{structured data} (P01, P10, P29, P53), and the emphasis on \textbf{routine tasks} (P01, P03, P06, P10, P51). 

Furthermore, some publications emphasise the \textbf{non-invasiveness} of RPA, meaning that RPA does not change the underlying application systems (P06, P46, P51).

In 2017, the IEEE Standards Association defined RPA as follows \citep{IEEE}: 
`A preconfigured software instance that uses business rules and predefined activity choreography to complete the autonomous execution of a combination of processes, activities, transactions, and tasks in one or more unrelated software systems to deliver a result or service with human exception management.'
This definition includes the aspects \textbf{software-based}, \textbf{rule-based}, and \textbf{non-invasive}. The other aspects, namely \textbf{mimics human behaviour} and are \textbf{routine tasks} with \textbf{structured data} are not covered. Moreover, this definition includes the goal of implementing RPA (\textbf{`deliver a result or service'}), and it emphasises that \textbf{humans are needed to handle exceptions}. Note that these two aspects are not addressed by any other definition. \\ 

\textbf{Differences of RPA to Related Technologies.} In the following, we analyse the differences between RPA and \textit{Robotic Desktop Automation (RDA)}, \textit{Intelligent/Cognitive RPA}, and \textit{BPM}. These technologies are the most frequently mentioned ones in the results of the SLR.

As major difference between RDA and RPA, \textbf{RDA does not have its own identity} and, therefore, acts via the IT infrastructure of its users with the same roles and authorisations, whereas RPA is working autonomously in the background on a central server structure (P40). Furthermore, \textbf{RDA is attended}, whereas RPA is unattended (P40).  Additionally, scripting and screen scraping are locally deployed from the user's desktop and can be seen as RDA, differing from RPA, which is enterprise-safe, meeting IT requirements such as security, scalability, auditability, and change management (P58). In P51, stand-alone automation includes macros, office program automation, and mouse/keyboard emulation.

Most publications distinguish between intelligent and cognitive automation. Intelligent or enhanced RPA, also called self-learning RPA, uses data to learn how a user interacts with the system and mimics these interactions including human judgement (P06, P19, P51). Machine learning and process mining techniques \citep{VanderAalst2011} are used to build knowledge of the process to better automate it (P51, P54).
Cognitive RPA, in turn, uses advanced machine learning and natural language processing to augment human intelligence and to learn performing tasks in a better way (P06, P43, P54). The main differences between rule-based automation and intelligent automation are summarised in Table~\ref{tab:diarpa}.

\begin{table}
\setcounter{table}{2}
\tbl{Differences between RPA and intelligent automation.}
{\begin{tabular}{llll} \toprule
Criterion & RPA & Intelligent Automation & Ref \\ \midrule
Degree of standardisation & high & low & P58 \\ 
Data & structured & unstructured & P51, P54 \\
Decisions & rule-based & knowledge/experience-based & P06, P19, P51 \\ 
Outcome & deterministic & probabilistic & P01, P54 \\ 
Exceptions & demand human intervention & trigger machine learning & P43, P54 \\ \bottomrule
\end{tabular}}
\label{tab:diarpa}
\end{table}

\begin{figure}
\centering
\includegraphics[width=\textwidth]{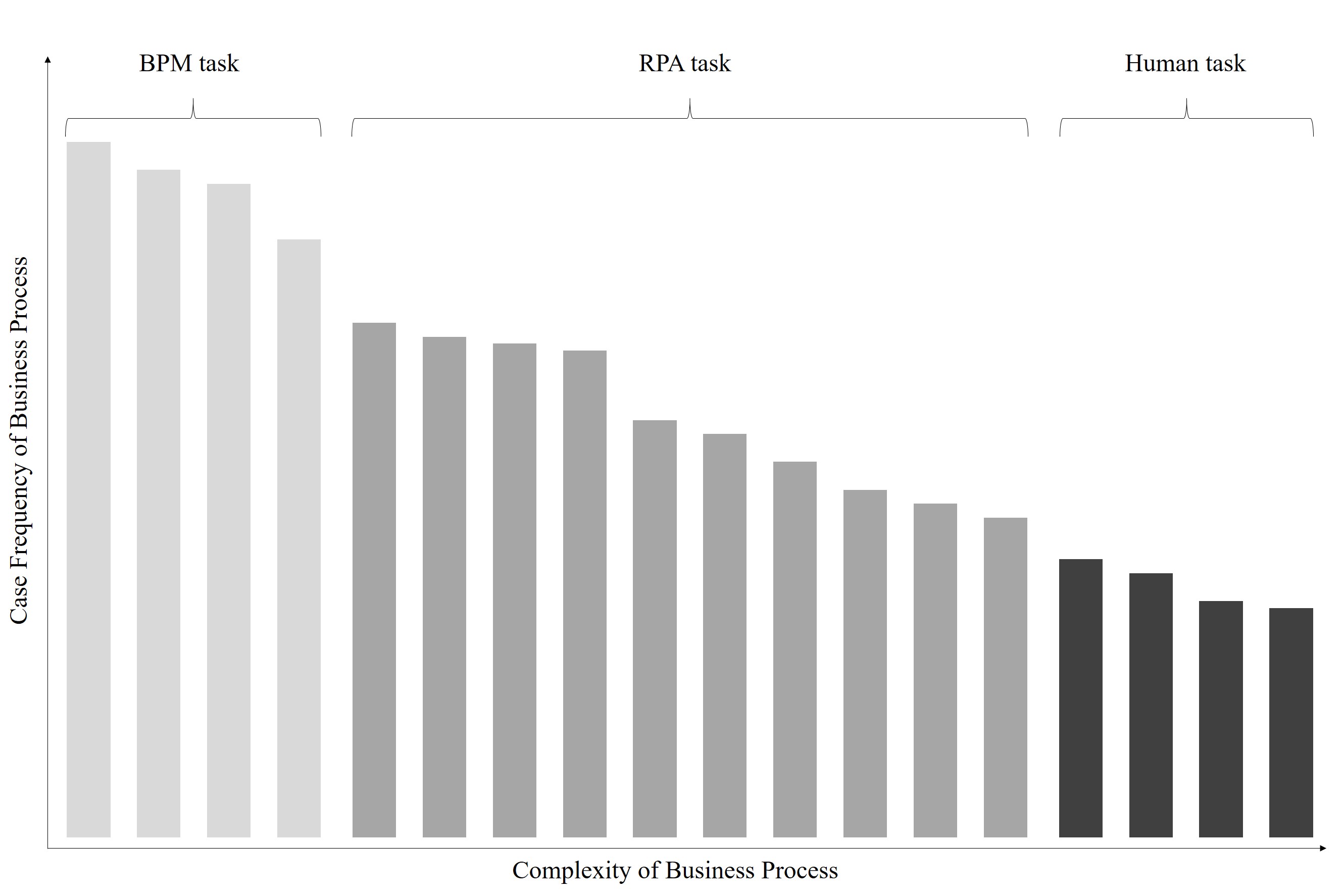}
\caption{Comparison of tasks suitable for BPM, for RPA, and tasks only Humans can do.}
\label{img:BPMvsRPA}
\end{figure}

Many publications emphasise the differences between RPA and BPM. Figure~\ref{img:BPMvsRPA} illustrates these differences graphically. The x-axis indicates the number of process variants (i.e., the complexity of the business process). The y-axis displays the case frequency of all process variants of the business process. The tasks on the left are best suited for BPM, the ones in the middle are candidate tasks for RPA, and the ones on the right can only be performed by humans (cf. Figure~\ref{img:BPMvsRPA}) (P56, P60). 

Table~\ref{tab:dbpmrpa} summarises the main differences between RPA and BPM.

\begin{table}
\setcounter{table}{3}
\tbl{Differences between RPA and BPM.}
{\begin{tabular}{lp{5cm}p{5cm}l} \toprule
Criterion & RPA & BPM & Ref \\ \midrule
Goal & automate existing processes to reduce human interaction & re-engineer processes to optimise them & P07 \\
General idea & change `where' work is done & change `how' work is done & P07, P44, P53 \\
Invasiveness & non-invasive, lightweight IT sitting on top of existing business applications & heavyweight IT interacting with business logic and creating new business applications & P01, P07, P46, P60 \\
Problems & privacy, security issues & high complexity, expensive & P44, P46, P53 \\ \bottomrule
\end{tabular}}
\label{tab:dbpmrpa}
\end{table}

To understand the difference between lightweight and heavyweight IT (Table~\ref{tab:dbpmrpa}, row 3), we summarise characteristics of suitable tasks for both types of automation. Lightweight IT automates tasks involving multiple systems and having a high volume, and provide a stable user interface (UI). Heavyweight IT automates tasks working in one system, the tasks have a very high volume, and are characterised by a stable back-end system architecture (P07).

P56 emphasises another differentiation: RPA versus \textit{Straight Through Processing (STP)}. STP refers to processes that can be performed without any human involvement, whereas RPA is an `outside-in' approach, which uses existing information systems and shall be robust to changes of these systems.

\subsection{RQ 2: Which business processes can be automated with RPA and which tools are used for automation?}
\label{ch:rq2}

\textbf{Process Selection Criteria.} The most frequently mentioned criterion in literature is \textbf{repetitiveness}, i.e., the process to be automated by robots shall have a high volume of transactions or a large number of process executions (P06, P10, P14, P28, P31, P50, P58, P59, P62, P63). Regarding the predictability of the process volumes, P06 states that processes with unpredictable peaks are suited for RPA implementations. However, P31 emphasises that the volumes should be predictable. 

Another important criterion concerns the \textbf{rule-based} character of the process. Consequently, the process to be automated shall be standardised, run in a stable environment, and only require limited exception handling (P02, P06, P14, P28, P31, P46, P62, P63).

The next criterion is to check whether the process requires \textbf{high manual efforts} and, thus, is prone to errors (P06, P14, P28, P51). Furthermore, digitisation gaps in processes might fulfil this criterion as they indicate the need for human work. P51 even states that `any activity that a person performs with mouse and keyboard can be carried out by a software robot.'

The \textbf{complexity} of the process itself, or as a result the complexity of its implementation, constitutes another important selection criterion. All publications agree that the lower the complexity, the better the process is suited for RPA (P17, P50, P58, P59).

Further, the \textbf{duration of process execution} can serve as a criterion. Processes to be automated shall have a high expenditure of time (P14, P17).

Additionally, the following criteria are mentioned by a few publications: The inputs and outputs are digital and structured (P10, P28, P46), the process only requires a limited number of human interventions, the process accesses multiple applications, the effects of a business failure are high (P14, P28), and the transaction has a great influence on the business (P14, P63). P06 proposes to choose processes for RPA automation, which are not a priority for the IT department. \\

\textbf{Use Cases.} Table~\ref{tab:bp} shows the 15 case studies, indicating in which \textit{Business Area} RPA was applied, which \textit{Business Process} was automated, and which \textit{Tool} was used for automation. 

The most present business areas is \textbf{Business Process Outsourcing (BPO)} (P01, P17, P30), followed by \textbf{Shared Services} (P32, P58), \textbf{Telecommunication} (P50, P59), and \textbf{Banking} (P40, P53). One case study was conducted in Digital Forensics (P03), Auditing (P10), Energy Supply (P31), Manufacturing (P47), Corporate Service Provider (P61), and Software Testing (P63) respectively.

Most automated processes are \textbf{swivel-chair processes}, i.e., `processes where humans take inputs from one set of systems (for example email), process those inputs using rules, and then enter the outputs into systems of record (for example Enterprise Resource Planning (ERP) systems)' (P60).

For ten case studies, the used automation tool was mentioned in the corresponding publication. Four used \textbf{Blue Prism} (P30, P53, P58, P59), two used \textbf{UiPath} (P03, P17), and one case study used Redwood (P32), Bluepond (P50), Workfusion (P63), and Roboplatform (P40) respectively. The latter is a self-made tool that was built in-house. P23 compares different automation tools, namely UiPath, Automation Anywhere, and Blue Prism based on criteria, e.g., openness of the platform, future scope or performance. Their recommendation is to use UiPath because it `triumphs all' (P23).

\begin{table}
\setcounter{table}{4}
\tbl{Business Area, Business Process, and Automation Tool for concrete Use Cases.}
{\begin{tabular}{llp{8.1cm}l} \toprule
Ref & Business Area & Business Process & Automation Tool\\ \midrule
P01 & BPO & Generate payment receipt & - \\
P03 & Digital Forensic & Search for keywords within Autopsy forensic software and import evidence files, process them and carry out image extraction in Griffeye forensic software & UiPath \\
P10 & Auditing & Collect data; copy it to template; filter, prepare, transfer it to database, and perform audit tests for loan testing & - \\
P17 & BPO & Update employee payment details and create new employment relationships & UiPath \\
P30 & BPO & Create and validate Premium Advice Notes & Blue Prism \\
P31 & Energy Supply & Resolve infeasible customer meter readings & - \\
P32 & Shared Services & Generate financial close & Redwood \\
P40 & Banking & Copy details of personal loan or current account from mainframe application to Excel & Roboplatform \\
P47 & Manufacturing & Master data management & - \\
P50 & Telecommunication & Bundle support tools for field service technician & Bluepond \\
P53 & Banking & Manage information interaction between bank and governmental institution & Blue Prism \\
P58 & Shared Services & Copy data from Excel to HRM System & Blue Prism \\
P59 & Telecommunication & Carry out SIM swaps and apply pre-calculated credit to account & Blue Prism\\
P61 & Corporate Service Provider & Generation of documents for annual compliance process and handle ad-hoc inquiries of customer & - \\
P63 & Software Testing & Schedule and control software testing by executing test scripts and validating the UI & Workfusion \\ \bottomrule
\end{tabular}}
\label{tab:bp}
\end{table}

\subsection{RQ 3: What are RPA effects?}

The answer to RQ 3 is divided into two aspects: the first one deals with the RPA effects on humans and their work life, whereas the second one deals with positive, controversially discussed, and negative effects on the company. 

As a positive effect of RPA on employees, the latter are \textbf{relieved from non-value adding tasks} and, consequently, they become more satisfied (P12, P17, P25, P33, P47, P51, P55, P58). New tasks and jobs for employees are proposed. One area concerns the development, testing, and monitoring of software robots (P02, P28, P32). Most publications mention that humans can \textbf{focus on cognitively more demanding tasks} (P13, P17, P32, P41), including activities that require judgement, interpretation, and assessment of results (P03, P10, P29, P54, P58). Furthermore, unstructured tasks (P29, P32, P54), creative tasks (P32), and tasks demanding for empathy and social interactions (P29, P58) are best suited for humans, e.g., to build relationships with the customer (P54). 

According to (P02, P13, P17, P18, P30, P55), employees \textbf{fear to lose their job}. They consider the robots as their competitors for their job (P02, P18) and are \textbf{afraid to learn the use of the new technology} (P13, P14). Hence, acceptance problems might arise (P18). P29 and P54 propose combined human robot teams, where each team member performs the task he or she can do best.  In P30, myths about RPA are demythologised, e.g., `RPA is only used to replace humans with technology'. In turn, P30 is refuted by the fact that more work can be done with the same number of people and humans are not replaced by technology. According to P61, staff reduction is one effect of RPA implementations. 

\begin{figure}
\centering
\includegraphics[width=\textwidth]{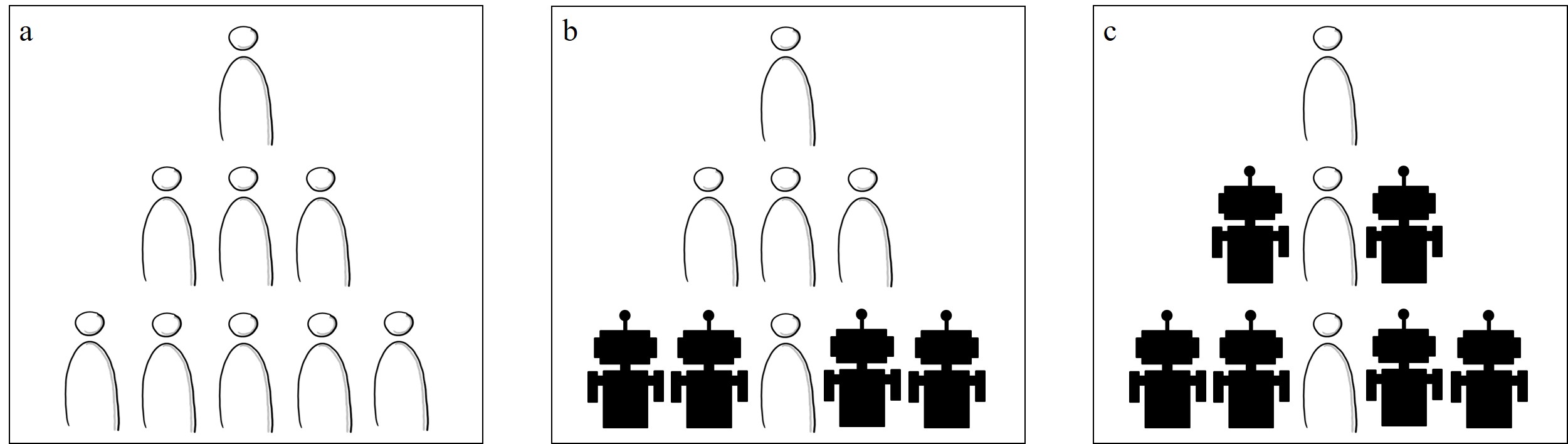}
\caption{Organisational Structure in a Company - Human Workforce is a a: Pyramid b: Diamond c: Pillar.}
\label{img:pyramid}
\end{figure}

According to (P14, P32, P54), there will be \textbf{less tasks for humans}, especially regarding low-level tasks not requiring any specific qualification. P11 and P12 emphasise that even knowledge workers are affected by lay-off due to RPA. On one hand, this has an impact on jobs in low-cost countries (P32). 
P30 proposes to automate offshore processes and keep them offshore, whereas P03 stresses that humans are needed to trigger the robot. On the other, organisational structures change. Nowadays, most companies are structured like a pyramid, having many less-skilled workers and fewer highly skilled workers (cf. Figure~\ref{img:pyramid}a). P32 predicts the change from that pyramid structure to a diamond structure meaning that employees at the bottom of the pyramid will be replaced by robots (cf. Figure~\ref{img:pyramid}b). P42 goes further and predicts that the pyramid structure will be replaced by a pillar structure regarding the human workforce (cf. Figure~\ref{img:pyramid}c). Robots will fill up the structure such that the overall organisation structure remains a pyramid. 

The positive RPA effects on the company, excluding human aspects, can be clustered in four categories:
\begin{itemize}
	\item \textbf{Speed:} Automated processes run faster and case duration becomes shorter (P01, P12, P14, P18, P28, P29, P35, P47, P50, P54, P55, P58, P61).
	\item \textbf{Availability:} Most RPA bots are available 24/7, and instant access is granted. Moreover, RPA is highly scalable to meet a varying intensity of demands (P01, P06, P14, P29, P33, P43, P50, P51, P59, P62).
	\item \textbf{Compliance:} Processes executed by a bot are highly transparent and documented in detail. Therefore, compliance is increased (P29, P32, P33, P35, P43, P51, P59).
	\item \textbf{Quality:} RPA eliminates human errors, improves accuracy and data quality, and leads to a higher customer satisfaction (P06, P12-P14, P28, P35, P43, P47, P50, P51, P54, P55, P58, P59, P62).
\end{itemize}

There are some effects of RPA projects that are controversially discussed. P43 and P46 criticise that RPA is \textbf{unable to make decisions}, P50 argues that it provides full transparency of all decisions. The latter means that if the bot fails, an employee still can perform the task manually. Section~\ref{ch:ai} discusses how AI is used to expand the limits of RPA, including decision making. The \textbf{costs} of RPA are another point of discussion: in P14 and P55 budget constraints are seen as a challenge to realise RPA projects, while many publications highlight its cheapness, cost reduction, and high return on investment (P03, P06, P13, P18, P31, P32, P33, P35, P46, P47, P50, P54, P59). P18 differs between the implementation and the maintenance:  The first is characterised by low costs, the latter can be costly and tedious. The \textbf{non-invasiveness} of RPA is seen differently: P03 and P46 criticise that RPA presumes an existing infrastructure and depends on the stability, availability, and performance of the systems. On the other, P54 considers the non-invasiveness as a benefit. P04 starts a discussion on possible RPA effects on enterprise architectures and argues that RPA might become invasive, i.e., RPA enables new work flows, requiring a modelling functionality in RPA systems, which contradicts the basic RPA idea. P46 emphasises that RPA is unable to adapt to a changing environment, whereas P02 and P62 notice that RPA is easily modifiable and flexible. 

Negative effects or limitations of RPA are seldom mentioned. Only P19 characterises RPA solutions as \textbf{workarounds} and P02 and P18 point out that RPA is a \textbf{temporary solution}. According to P03, there are software platforms, e.g., special forensic software, which are not compatible with current RPA solutions. Furthermore, P18 criticises that know-how and skills are required, and RPA solutions are not robust in respect to evolving user interfaces. P28 adds that RPA implementations require greater IT involvement than initially thought.

\subsection{RQ 4: Are there methods for improving the implementation of RPA projects?}

\begin{figure}
\centering
\includegraphics[width=\textwidth]{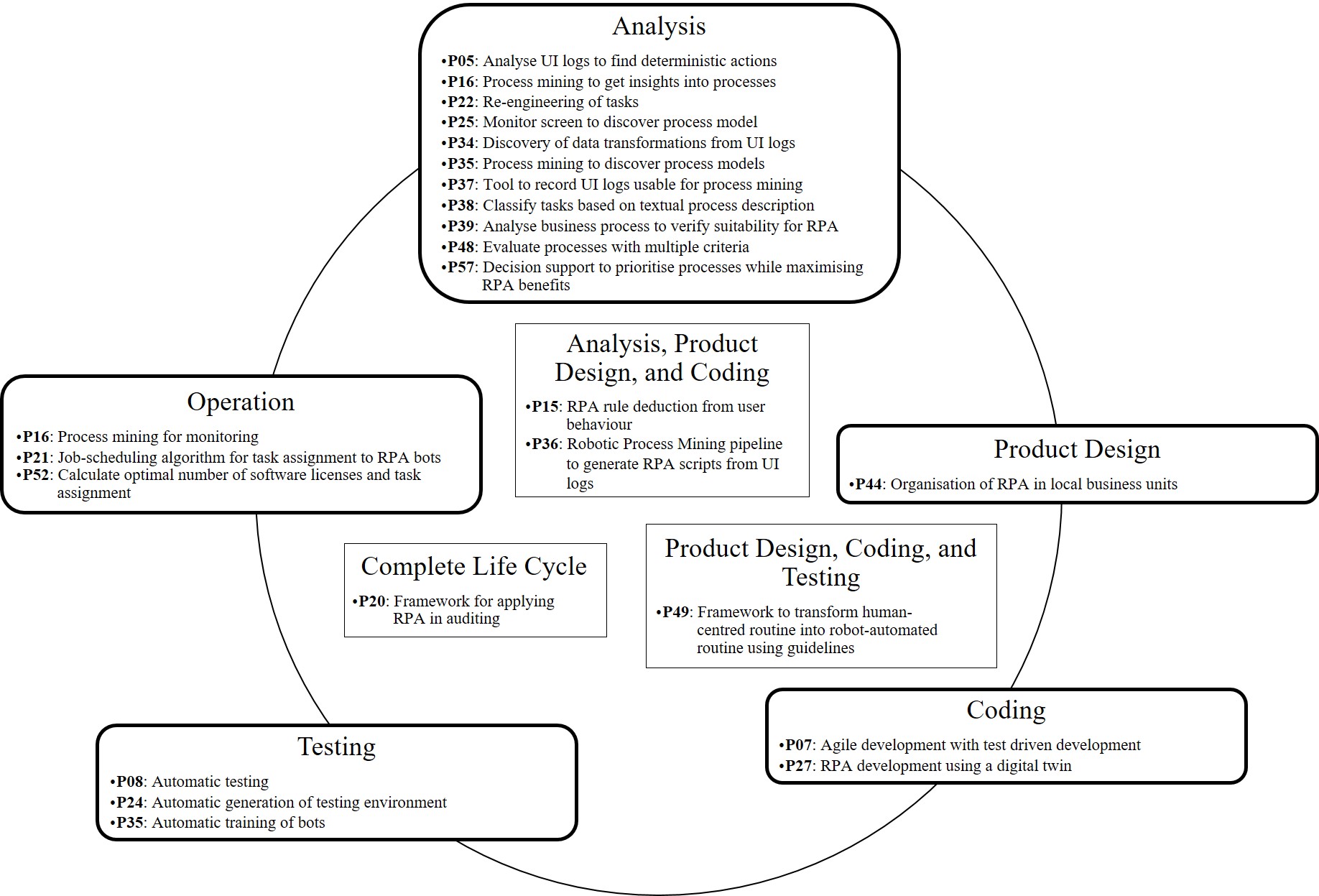}
\caption{SDLC annotated with Method Paper.}
\label{img:Methodpaper}
\end{figure}

To analyse the publications that introduce methods for RPA projects, we oriented ourselves on the software development life cycle (SDLC)\citep{Royce1987}. We assigned the methods to the corresponding stage in the life cycle for the sake of better illustration (cf. Figure~\ref{img:Methodpaper}). In the following, the methods are described shortly. 

\textbf{Analysis Stage.} The approaches to improve the Analysis Stage are roughly clustered into three areas: \textbf{process insights}, \textbf{process standardisation}, and \textbf{process selection}.

P16 uses process mining to get insights into the process, e.g., its automation rate. In P38, textual process descriptions are used to classify the tasks into the categories \textit{Manual Task}, \textit{User Task}, and \textit{Automated Task}. The goal is to automatically detect tasks suited for RPA. To achieve this goal, P38 uses feature computation for prediction and a Support Vector Machine (SVM) to classify the process descriptions based on the features. 

The aim of P35 is to develop a new process mining technique, which can deal with RPA and automatically discover process models. The approach is to discover constraints within an event log, extract corresponding feature vectors, and label constraint violations. P35 uses clustering methods to identify correlations between activation and target payloads. In a subsequent publication of the same authors, i.e. P37, a tool (`Action Logger') is developed, which records UI logs that can directly serve as inputs to process mining tools and contain information relevant for RPA implementation. P34, again by the same authors, develops an idea how to discover data transformations from UI logs.

In P16, process mining is used to standardise existing processes. Another standardisation technique is proposed in P22, which emphasises the importance of not automating the as-is process, but to optimise it before. Thus, the authors propose a framework for process re-engineering. 

The most difficult task in the analysis stage is to select the process to automate. Different approaches are proposed: P16 sticks to process mining for prioritising activities. P25 also uses process mining to discover processes, with a method focusing on creating event logs from screen monitoring data. P05 analyses UI logs to discover deterministic actions. As basic idea, `a routine is automatable if its first action is always triggered when a condition is met [...] and the value of each parameter of each action can be computed from the values of parameters of previous actions' (P05). 

P39 develops a four-step method to analyse a business process based on its criteria (cf. Section~\ref{ch:rq2}): first, to be eligible for RPA, the process has to be mature and standardised (Step 1). Step 2 assesses the RPA potential of the process based on human interaction with software and its rule-based nature. Step 3 evaluates the RPA relevance based on the volume of transactions and the degree of complexity of the process. Finally, based on Steps 2 and 3, the process is classified. P39 recommends to select processes with high relevance and high potential. In turn, P48 follows a similar approach and develops a multi-criteria process evaluation model, which assesses the technical feasibility and business potential criteria to find suitable business processes for RPA. The technical criteria include the degree of rule-basedness, human intervention, digitalisation, and the structuredness of data. The potential criteria evaluate labour intensity, the number of systems involved, the number of process exceptions, the number of process steps, current costs, and process maturity. P57 proposes a method to prioritise processes while maximising RPA benefits. Based on different indicators of the process, i.e., execution frequency, execution time, degree of standardisation, stability(i.e., small number of exceptions in the process) failure rate, and automation rate, the automation potential of the process is assessed. Furthermore, the profitability of process automation is measured through fixed and variable costs of human labour and fixed and variable costs of RPA. Finally, P57 maximises the economic value and provides recommendations to support the decision of selecting appropriate processes for RPA initiatives.

\textbf{Product Design Stage.} P44 highlights advantages and challenges of organising RPA in local business units. On the positive side, enthusiasm for digitalisation and local ownership are built. On the other, there is a lack of control mechanisms and end-to-end process views. P44 proposes to loosely couple the IT department and the RPA team. 

\textbf{Coding Stage.} P07 suggests a method for implementing RPA projects in an agile way: instead of documenting a process completely with clicks and text-based description, the users record themselves when performing the task and stores the video in the backlog. The developer creates a test case for this video and checks whether the current solution passes the test (Test Driven Development). If not, he modifies the RPA solution until the test case is fulfilled. Then, he moves on to the next video.
P27 proposes the use of digital twins for RPA development. A digital twin is, in this context, a virtual shadow of an IT system. The idea allows developing RPA externally without having access to the real system.

\textbf{Testing Stage.} P35 has the vision to develop a method `to automatically train the RPA bots'. Research has not progressed far enough. P08 proposes a method for automated testing in RPA projects, which has been tested with a prototype. The approach is to modify the RPA life cycle. Compared to the life cycle model depicted in Figure~\ref{img:Methodpaper}, the third stage is called development and not coding, operation is named monitoring, and a fourth stage, i.e., deployment, is inserted before the testing phase. The modified life cycle not only includes design in the second stage, but test environment construction as well. During development, automatic testing can be performed serving as new input for the analysis phase. P24 extends the approach of P08 by providing technical details on test cases and the algorithm as well as by evaluating the approach of automatically generating a testing environment.

\textbf{Operation Stage.} P16 mentions that process mining can be used to monitor the results of an RPA project.
P21 proposes a middle ware system for controlling the execution of multiple RPA bots. The system includes a job-scheduling algorithm to efficiently distribute multiple tasks among available bots.
In turn, P52 solves an optimisation problem to determine the optimal number of required bots while minimising costs. Then, the optimal task assignment among the bots is solved.

Some publications cannot be assigned to solely one stage and are, therefore, placed in the middle of Figure~\ref{img:Methodpaper}. P15 and P36 cover the first three stages, i.e., Analysis, Product Design, and Coding. 
To be more precise, P15 presents an end-to-end approach that allows deducing RPA rules from user behaviour. The idea is based on the Form-to-Rule approach: First, tasks of the user are identified by observing interactions with systems and identifying forms used within the systems. Second, rules are deduced from relations between the different tasks. Third, RPA is implemented based on those rules.
P36 combines the approaches presented in P34, P35, and P37 and proposes a Robotic Process Mining pipeline. After recording UI logs, noise filtering, segmentation, and simplification steps are applied to identify candidate routines. In these routines, executable (sub)routines are discovered and compiled to obtain RPA scripts. P36 emphasises that there are still many challenges to successfully apply the proposed pipeline.

Stages Product Design, Coding, and Testing are addressed by P49. A framework is developed to transform a human-centred routine into a robot-automated one. The framework of routine automation can be empirically applied to different areas, including RPA, and provides implementation guidelines. 

One publication, i.e., P20, addresses the complete life cycle of RPA and proposes a framework to introduce RPA in auditing. The first stage is the process selection based on the evaluation of different criteria, e.g., RPA criteria, process complexity, and data compatibility. Second, the process is modified, e.g., considering data standardisation. In a third step, the process is implemented and, finally, evaluated and operated. The last step consists of evaluating effectiveness, assessing detection risk (i.e. the risk that auditing `will not detect a misstatement' (P20)), and monitoring the RPA operations.

\subsection{RQ 5: Is AI used in combination with RPA?}
\label{ch:ai}

To get an idea whether AI is already used in combination with RPA, our insights from literature are summarised in the following. 

Some works \textbf{briefly mention} the use of AI and its potentials. P40 and P43 state that with AI it becomes possible to understand semi-structured data. P56, in turn, emphasises that AI helps interpreting changing user interfaces and improving the robustness of RPA solutions. Using chat bots, P43 presumes that the interaction between humans and computer systems becomes facilitated. 

First \textbf{AI-based applications} have emerged in the RPA field: P26 presents a Cognitive Automation Robots Platform, which is able to understand data, generate insights, and use the latter as learning experiences. P33 uses the cognitive virtual agent `Amelia', which understands chat messages. In P19, a cognitive RPA prototype is presented. It can automatically identify, extract, and process data. Once the classification model is trained (for details see P19, pp. 68-69), new unseen documents are classified and relevant objects, e.g., address fields, are detected and extracted. 

We discovered four publications that combine AI with RPA in greater detail. P41 provides building blocks for intelligent process automation by explaining and providing implementations on how to extract intent from audio, classify emails, detect anomalies, find cross correlations in time series, and understand traffic patterns. P62 describes how machine learning methods contribute to further improve RPA, e.g., using image processing to scan letters or invoices or using classification algorithms to label documents.

The task of \textbf{classifying emails} correctly is picked up by several publications: P45 proposes the use of an SVM and a Text Rank Algorithm to read emails and to automatically process them. P09 develops an algorithm, named Sure-Tree, for email classification, which produces a minimum of false positives to ensure that an incorrect action is never triggered.

\section{Deriving a Framework for Analysing and Comparing RPA Publications}
\label{ch:ANCOPUR}

This section synthesises the results obtained by the SLR. More precisely, we present a framework for \textbf{An}alysing and \textbf{Co}mparing existing as well as upcoming \textbf{Pu}blications in \textbf{R}PA (ANCOPUR for short). ANCOPUR gathers the results along the defined research questions (cf. Section~\ref{ch:rq}). 

Table~\ref{tab:fw} depicts the schema of our ANCOPUR framework: The first column, shows the main aspect for comparison, e.g., definition, process selection criteria, or use case. In the following columns the aspect gets detailed. The publication can be assigned to several rows depending on the aspects it covers. 

If a new feature is found, it can be added to ANCOPUR as well. To demonstrate its usefulness and applicability, all 63 publications from the SLR are categorised with ANCOPUR. Note that this facilitates the comparison of any new publication with existing knowledge. 

We illustrate and explain ANCOPUR by assigning P17 exemplary to it. This publication was part of the results of the SLR. Furthermore, a publication randomly excluded in the results of the SLR, is assigned to the framework to evaluate it \citep{Flechsig2019}.

Publication P17 is read to detect information depositing on the first column of the ANCOPUR framework. We discover the aspects \textit{Process Selection Criteria}, \textit{Effects}, and \textit{Use Case}. 
Concerning the criteria for processes to be automated, P17 emphasises `1) the processes should be simple enough so that the robots could be implemented quickly and 2) improved process efficiency resulting from RPA implementation should be clearly visible.' \citep{Hallikainen2018} Therefore, the processes are selected depending on their \textbf{complexity} and the \textbf{duration of process execution}. Regarding ANCOPUR, P17 is indeed assigned to those two rows in the Process Selection Criteria column. 
Regarding the case study, the aspects \textit{Business Area}, \textit{Business Process}, and \textit{Automation Tool} are all covered by P17: the general business area is BPO and the concrete processes are `1) new employment relationships and 2) changes in employee payment details', which are both swivel-chair processes. UiPath was used for automating the business processes. Therefore, P17 is added as reference to rows \textbf{BPO}, \textbf{swivel-chair process}, and \textbf{UiPath} in the use case section of ANCOPUR. 
The following wording is found for RPA effects: `there were some fears about losing jobs [...] people would no longer have to carry out the boring work and could concentrate on more interesting tasks' (P17). The first statement expresses a negative effect on humans and is assigned to \textbf{fear to lose the job} in ANCOPUR. The second statement describes positive effects on humans and covers both aspects in ANCOPUR, namely \textbf{relieved from non-value adding tasks} and \textbf{focus on cognitively more demanding tasks}. 

To evaluate ANCOPUR, we assign the work presented by \citep{Flechsig2019} to it. For \textit{Definition}, \textit{Use Case}, \textit{Effect}, and \textit{Combination with AI}, no new aspects or no information are found at all.
Concerning \textit{Differences of RPA to Related Technologies}, \textit{BPM} is compared to RPA. All aspects in ANCOPUR are covered, only the formulations differ a bit, e.g., `Redesign of extensive processes with high strategic relevance and added value' \citep{Flechsig2019} is assigned to the row \textbf{changes `how' work is done}.
Regarding \textit{Process Selection Criteria}, we find several questions that aim to find processes suitable for an RPA implementation \cite[p. 111]{Flechsig2019}. These criteria include \textbf{repetitive}, \textbf{rule-based}, \textbf{duration of process execution}, and \textbf{high effects of business failure}. Therefore, \citep{Flechsig2019} can be added to the corresponding rows in the ANCOPUR framework. Additionally, \citep{Flechsig2019} proposes choosing processes relevant for \textbf{compliance}, an aspect not considered yet. ANCOPUR can be expanded by this process selection criterion aspect. 
\citep{Flechsig2019} suggests a method for combining BPM and RPA, which can be assigned to the \textit{Product Design Stage} with a new row, namely \textbf{combination of BPM and RPA}. The idea is to have a common Analysis Stage for BPM and RPA projects as well as to decide in the Product Design Stage whether to implement a BPM or an RPA solution. 

\begin{sidewaystable}
\setcounter{table}{5}
\tbl{Framework ANCOPUR.}
{\begin{tabular}{@{}llllp{7cm}} \toprule
\multicolumn{4}{c}{ANCOPUR} & Ref \\ \midrule
\multirow{8}{*}{\parbox{1.6cm}{Definition}} & \multicolumn{3}{l}{software-based solution} & \citep{IEEE}, P02, P33, P40, P43, P51, P58, P59, P60 \\ \cline{2-5}
& \multicolumn{3}{l}{mimics human behaviour} & P06, P40, P43, P45, P51, P53, P56, P59, P60 \\ \cline{2-5}
& \multirow{3}{*}{\parbox{3.1cm}{characteristics of the automated process}} & \multicolumn{2}{l}{rule-based} & \citep{IEEE}, P01, P10, P29, P45 \\ \cline{3-5}
& & \multicolumn{2}{l}{structured data} & P01, P10, P29, P53 \\ \cline{3-5}
& & \multicolumn{2}{l}{routine tasks} & P01, P03, P06, P10, P51 \\ \cline{2-5}
& \multicolumn{3}{l}{non-invasive} & \citep{IEEE}, P06, P46, P51 \\ \cline{2-5}
& \parbox{3.1cm}{goal of implementation} & \multicolumn{2}{l}{deliver result/service} & \citep{IEEE} \\ \cline{2-5}
& \multicolumn{3}{l}{humans handle exceptions} & \citep{IEEE} \\ 
\hline
\multirow{12}{*}{\parbox{1.6cm}{Differences of RPA to Related Technology}} & \multirow{2}{*}{\parbox{3.1cm}{RDA}} & \multicolumn{2}{l}{no own identity} & P40 \\ \cline{3-5}
& & \multicolumn{2}{l}{attended automation} & P40 \\ \cline{2-5}
& \multirow{5}{*}{\parbox{3.1cm}{Intelligent/Cognitive Automation}} & \multicolumn{2}{l}{low degree of standardisation} & P58 \\ \cline{3-5}
& & \multicolumn{2}{l}{unstructured data} & P51, P54 \\ \cline{3-5}
& & \multicolumn{2}{l}{knowledge/experience-based decisions} & P06, P19, P51 \\ \cline{3-5}
& & \multicolumn{2}{l}{probabilistic outcome} & P01, P54 \\ \cline{3-5}
& & \multicolumn{2}{l}{exceptions trigger machine learning} & P43, P54 \\ \cline{2-5}
& \multirow{4}{*}{\parbox{3.1cm}{BPM}} & \multicolumn{2}{l}{re-engineer processes to optimise them} & P07 \\ \cline{3-5}
& & \multicolumn{2}{l}{changes `how' work is done} & P07, P44, P53 \\ \cline{3-5}
& & \multicolumn{2}{l}{creates new business applications} & P01, P07, P46, P60 \\ \cline{3-5}
& & \multicolumn{2}{l}{highly complex and expensive} & P44, P46, P53 \\ \cline{2-5}
& \multicolumn{3}{l}{STP} & P56 \\ 
\hline
\multirow{9}{*}{\parbox{1.6cm}{Process Selection Criteria}} & \multicolumn{3}{l}{repetitive} & P06, P10, P14, P28, P31, P50, P58, P59, P62, P63 \\ \cline{2-5}
& \multicolumn{3}{l}{rule-based} & P02, P06, P14, P28, P31, P46, P62, P63 \\ \cline{2-5}
& \multicolumn{3}{l}{high manual effort} & P06, P14, P28, P51 \\ \cline{2-5}
& \multicolumn{3}{l}{complexity} & P17, P50, P58, P59 \\ \cline{2-5}
& \multicolumn{3}{l}{duration of process execution} & P14, P17 \\ \cline{2-5}
& \multicolumn{3}{l}{digital and structured inputs and outputs} & P10, P28, P46 \\ \cline{2-5}
& \multicolumn{3}{l}{limited number of human intervention} & P14 \\ \cline{2-5}
& \multicolumn{3}{l}{access to multiple applications} & P14, P28 \\ \cline{2-5}
& \multicolumn{3}{l}{high effects of business failure} & P14, P63 \\ 
\hline
\multirow{17}{*}{\parbox{1.6cm}{Use Case}} & \multirow{10}{*}{\parbox{3.1cm}{Business Area}} &  \multicolumn{2}{l}{BPO} & P01, P17, P30 \\ \cline{3-5}
& &  \multicolumn{2}{l}{Shared Services} & P32, P58 \\ \cline{3-5}
& &  \multicolumn{2}{l}{Telecommunication} & P50, P59 \\ \cline{3-5}
& &  \multicolumn{2}{l}{Banking} & P40, P53 \\ \cline{3-5}
& &  \multicolumn{2}{l}{Digital Forensics} & P03\\ \cline{3-5}
& &  \multicolumn{2}{l}{Auditing} & P10 \\ \cline{3-5}
& &  \multicolumn{2}{l}{Energy Supplier} & P31\\ \cline{3-5}
& &  \multicolumn{2}{l}{Manufacturing} & P47 \\ \cline{3-5}
& &  \multicolumn{2}{l}{Corporate Service Provider} & P61 \\ \cline{3-5}
& &  \multicolumn{2}{l}{Software Testing} & P63 \\ \cline{2-5}
& \parbox{3.1cm}{Business Process} & \multicolumn{2}{l}{swivel-chair process} & P01, P03, P10, P17, P30-P32, P40, P47, P50, P53, P58, P59, P61, P63\\ \cline{2-5}
& \multirow{6}{*}{\parbox{3.1cm}{Automation Tool}} & \multicolumn{2}{l}{Blue Prism} & P30, P53, P58, P59 \\ \cline{3-5}
& &  \multicolumn{2}{l}{UiPath} & P03, P17 \\ \cline{3-5}
& &  \multicolumn{2}{l}{Redwood} & P32 \\ \cline{3-5}
& &  \multicolumn{2}{l}{Bluepond} & P50 \\ \cline{3-5}
& &  \multicolumn{2}{l}{Workfusion} & P63 \\ \cline{3-5}
& &  \multicolumn{2}{l}{Roboplatform} & P40 \\ 
\bottomrule
\end{tabular} }
\label{tab:fw}
\end{sidewaystable}

\begin{sidewaystable}
\setcounter{table}{6}
\tbl{Framework ANCOPUR - Continuation.}
{\begin{tabular}{@{}llllp{7cm}} \toprule
\multirow{14}{*}{\parbox{1.6cm}{Effect}} & \multirow{5}{*}{\parbox{3.1cm}{on humans and future work life}} & \multirow{2}{*}{\parbox{2.2cm}{positive}} & relieved from non-value adding tasks & P17, P25, P33, P47, P51, P55, P58 \\ \cline{4-5}
& & & focus on cognitively more demanding tasks & P02, P03, P10, P12, P13, P17, P28, P29, P32, P41, P54, P58 \\ \cline{3-5}
& & \multirow{3}{*}{\parbox{2.2cm}{negative}} & fear to lose the job & P02, P13, P17, P18, P30, P55 \\ \cline{4-5}
& & & afraid to learn new technology, acceptance problems & P13, P14, P18 \\ \cline{4-5}
& & & less tasks, lay-off & P11, P12, P14, P32, P42, P54, P61 \\ \cline{2-5}
& \multirow{9}{*}{\parbox{3.1cm}{on company}} & \multirow{4}{*}{\parbox{2.2cm}{positive}} & speed & P01, P12, P14, P18, P28, P29, P35, P47, P50, P54, P55, P58, P61 \\ \cline{4-5}
& & & availability & P01, P06, P14, P29, P33, P43, P50, P51, P59, P62 \\ \cline{4-5}
& & & compliance & P29, P32, P33, P35, P43, P51, P59 \\ \cline{4-5}
& & & quality & P06, P12-P14, P28, P35, P43, P47, P50, P51, P54, P55, P58, P59, P62 \\ \cline{3-5}
& & \multirow{3}{*}{\parbox{2.2cm}{controversially discussed}} & inability to make decisions & P43, P46, P50 \\ \cline{4-5}
& & & costs & P03, P06, P13, P14, P18, P31-P33, P35, P46, P47, P50, P54, P55, P59 \\ \cline{4-5}
& & & non-invasiveness & P03, P46, P54 \\ \cline{3-5}
& & \multirow{2}{*}{\parbox{2.2cm}{negative}} & workaround, temporary solution & P02, P18, P19 \\ \cline{4-5}
& & & incompatibility of software with RPA solutions & P03 \\ \hline
\multirow{20}{*}{\parbox{1.6cm}{Method}} & \multirow{8}{*}{\parbox{3.1cm}{Analysis Stage}} & \multirow{3}{*}{\parbox{2.2cm}{process insights}} & automation rate & P16 \\ \cline{4-5}
& & & classify tasks based on textual process description & P38 \\ \cline{4-5}
& & & discover process models & P25, P34, P35, P37 \\ \cline{3-5}
& & \parbox[c][0.8cm]{2.2cm}{process standardisation} & framework for task re-engineering & P22 \\ \cline{3-5}
& & \multirow{4}{*}{\parbox{2.2cm}{process selection}} & multi-criteria process evaluation model & P48 \\ \cline{4-5}
& & & analyse business processes based on criteria & P39 \\ \cline{4-5}
& & & prioritise business process, maximise benefits & P57 \\ \cline{4-5}
& & & analyse UI logs to find deterministic actions & P05 \\ \cline{2-5}
& \parbox{3.1cm}{Product Design Stage} & \multicolumn{2}{l}{organise RPA in local business units} & P44\\ \cline{2-5}
& \multirow{2}{*}{\parbox{3.1cm}{Coding Stage}} & \multicolumn{2}{l}{agile development} & P07\\ \cline{3-5}
& & \multicolumn{2}{l}{development with digital twin} & P27 \\ \cline{2-5}
& \multirow{2}{*}{\parbox{3.1cm}{Testing Stage}} & \multicolumn{2}{l}{automatic testing} & P08, P24 \\ \cline{3-5}
& & \multicolumn{2}{l}{automatic training} & P35 \\ \cline{2-5}
& \multirow{3}{*}{\parbox{3.1cm}{Operation Stage}} & \multicolumn{2}{l}{process mining to monitor results} & P16\\ \cline{3-5}
& & \multicolumn{2}{l}{algorithm for job-scheduling and task assignment} & P21 \\ \cline{3-5}
& & \multicolumn{2}{l}{optimal number of licences and task assignment} & P52 \\ \cline{2-5}
& \multirow{4}{*}{\parbox{3.1cm}{Several Stages}} & \multicolumn{2}{l}{RPA rule deduction from user behaviour} & P15 \\ \cline{3-5}
& & \multicolumn{2}{l}{generate RPA scripts from UI logs} & P36 \\ \cline{3-5}
& & \multicolumn{2}{l}{\parbox[c][0.8cm][c]{7.5cm}{framework to transform human-centred into robot-automated routine}} & P49 \\ \cline{3-5}
& & \multicolumn{2}{l}{framework for RPA in auditing} & P20 \\ \cline{3-5}
\hline
\multirow{5}{*}{\parbox{1.6cm}{Combination with AI}} & \multicolumn{3}{l}{briefly mention AI} & P40, P43, P56\\ \cline{2-5}
& \multicolumn{3}{l}{present prototype} & P19, P26, P33 \\ \cline{2-5}
& \multicolumn{3}{l}{machine learning methods} & P41, P62 \\ \cline{2-5}
& \multirow{2}{*}{\parbox{3.1cm}{classify emails}} & \multicolumn{2}{l}{SVM, Text Rank} & P45 \\ \cline{3-5}
& & \multicolumn{2}{l}{Sure-Tree} & P09 \\ 
\bottomrule
\end{tabular} }
\label{tab:fw2}
\end{sidewaystable}

Altogether, ANCOPUR uses criteria and sub-criteria to classify RPA publications. The framework is useful for systematically analysing, assessing, and comparing existing as well as upcoming RPA works.

\section{Related Work}
\label{ch:relatedwork}

Scientific works on RPA have been analysed in Section~\ref{ch:results}. This section gives a short overview of other literature research approaches highlighting the differences to our work.

\citep{Gotthardt2019} examines the current state of RPA as well as fundamental challenges in accounting and auditing. For this purpose, a literature review, interview results, and case studies are presented to summarise key factors. Unlike our work, no SLR is presented. Instead, \citep{Gotthardt2019} follow a domain-specific approach by focusing on accounting and auditing, with a special emphasis on the role of AI. 

A systematic mapping study is conducted in \citep{Enriquez2020} to analyse the current state-of-the-art of RPA. The main focus is to evaluate 14 commercial RPA tools regarding the coverage of 48 functionalities mapped to RPA life cycle phases. As major result, the Operation phase is covered by over 80\% of the RPA tools, whereas support for the Analysis phase is below 15\%.

An SLR is presented in \citep{Riedl2019}. Its aim is to derive an evaluation model to identify business processes, which in parts or entirely can be subjected to RPA. The main focus of this SLR is to derive selection criteria for assessing the RPA suitability of business processes as well as to develop a corresponding evaluation method. \citep{Riedl2019} apply the SLR method described in \citep{Kitchenham2004}. However, research questions, search strings, data sources, inclusion and exclusion criteria, and data analysis differ from the ones described in our work. Only 25 scientific research articles, case studies, and professional reports are considered, compared to the 63 in our SLR. Moreover, the results are differently clustered due to the focus on different research questions. 

\citep{Guner2020} presents a literature review on RPA cases to answer the question how RPA as a routine capability advances BPM practices. The results show that RPA, as a routine capability, advances practices at individual, organisational, and social levels.

\citep{Santos2019} provide an approach to evaluate RPA development in business organisations and industries. A conceptual model on relationships between RPA topics, identified in a literature review, is presented. The model consists of three steps, i.e., definition of strategic goals, process assessment, and tactical evaluation and factors for a successful RPA implementation. Influencing factors include benefits, disadvantages, selection criteria, future challenges, and future opportunities.

\citep{Ivancic2019} presents another SLR on RPA. Some of the research questions in \citep{Ivancic2019} sound similar to ours, e.g., `How is RPA defined (RQ2-1)' reads like RQ 1 (cf. Section~\ref{ch:rq}). Through examining search string, data sources, and inclusion as well as exclusion criteria, the differences become visible. The search results in 27 publications compared to 63 in our SLR. Definition and benefits of RPA, and differences to BPM are shortly mentioned, whereas our paper goes into detail and reveals many aspects undiscovered by previous SLRs. Tools used for automation (RQ 2), effects (RQ 3), methods (RQ 4), and the combination with AI (RQ 5) are completely ignored by this SLR. Furthermore, no framework utilising SLR results for assessing and comparing newly upcoming works has been developed. 

\citep{Syed2020} identifies contemporary RPA-related themes and challenges for future research by presenting an SLR. The first two research questions overlap slightly with RQ 1 and RQ 3 as presented in this article. However, \citep{Syed2020} focuses on the description of RPA readiness/RPA maturity in literature, the potential of RPA, an effective RPA methodology, and current and future technologies for RPA. In contrast, our paper emphasises differences between RPA and related technologies, methods for improving the implementation of RPA projects, and the combination of RPA with AI. \citep{Syed2020} uses the results of the SLR to highlight key research challenges for future RPA research, whereas we derive a framework for evaluating and comparing RPA publications in a structured way. 

Hence, to the best of our knowledge, no other publication addresses the problems presented in Section~\ref{ch:ps}.

\section{Discussion}
\label{ch:discussion}

The presented results enable us to answer the five research questions. In the following, the results are discussed and interpreted along the seven discovered \textbf{thematic clusters}.
More precisely, we identified, categorised, and analysed 63 publications belonging to the following seven clusters: RPA Definition, Differences of RPA to Related Technologies, Process Selection Criteria, RPA Use Cases, RPA Effects, RPA Project Methods, and Combination of RPA with AI.
As main result we obtain the ANCOPUR framework, which enables a structured overview of the SLR results. More specifically, the framework provides a fast and easy way to identify and categorise publications in the RPA area. In particular, comparing new works with existing knowledge becomes much simpler and more structured. Moreover, ANCOPUR can be easily expanded. If new publications reveal unconsidered aspects, those can be added to evolve the framework and keep it up to date. In detail:

\begin{itemize}
\item[1.] \textbf{RPA Definitions.} It is emphasised that RPA is a software-based solution mimicking human behaviour. These aspects are important to indicate the difference of RPA to hardware bots.
\item[2.] \textbf{Differences of RPA to Related Technologies.} Most papers emphasise the differences between RPA and Intelligent Automation as well as between RPA and BPM.
\item[3.] \textbf{Process Selection Criteria.} Best suited for an RPA automation are repetitive, rule-based and complex business processes demanding for high manual efforts.
\item[4.] \textbf{Use Cases.} The majority of use cases stem from business areas such as BPO and Shared Services. Note that this is reasonable as those areas possess many repetitive, rule-based business process as, for example \textit{generation of payment receipt} \citep{Aguirre2017}. Anyway, it would be interesting to encounter more RPA projects in knowledge-intensive business areas, e.g., in research and development or in healthcare. Furthermore, the literature covers only successful RPA projects, leaving room for further research on failed projects. Concerning the RPA tools used in the case studies, Blue Prism and UiPath are dominating. According to \citep{ZGartner2019}, however, there are other tools that should be considered: Automation Anywhere, EdgeVerve Systems, NICE, Workfusion, Pegasystems, and Another Monday. The application of the different tools to one concrete use case as well as tool performance should be compared in further research studies.
\item[5.] \textbf{RPA Effects.} The positive effects are widely discussed in literature. Only a minority is critical towards RPA. One reason can be the novelty of RPA (cf. Figure~\ref{img:dist} in Section~\ref{ch:results}), due to which the technology is hyped and negative effects do not want to be seen. It is emphasised that employees are relieved from non-value adding tasks, and instead can focus on cognitively more demanding tasks. Finally, business processes become faster, better available, more compliant, and improved in quality.
\item[6.] \textbf{RPA Project Methods.} Most methods for improving the implementation of RPA were published in 2019 and 2020 (16 of 22 paper). The vast majority of methods tries to improve the analysis stage, only some publications address the other life cycle stages. The analysis stage is the one that differs mostly from other software development projects. Product design, coding, and testing are not differing that much when either implementing an RPA project or any other software project. We expect that more publications dealing with analysis will appear as well as methods to fully automate the detection of RPA-suitable processes. Furthermore, the operation stage should be addressed, e.g., it should be monitored whether the bots are accepted or employees fear to lose their job and, therefore, refuse the use of the bots.
\item[7.] \textbf{Combination of RPA with AI.} The use of AI in the context of RPA is still at a very early stage. Six publications deal with this combination from a general point of view and emphasise that it might create a big impact. Only four concrete use cases are discovered, the majority focuses on the problem of classifying emails correctly. While the use cases are still scientific in nature, it is interesting to see more industry-driven approaches and projects. The publications are from the last years only, therefore, we hope for more research in the coming years. 
\end{itemize}

In general, research on RPA is still at its beginning. Though being increasingly present in industry, scientific works on this topic are rather scarce and mainly consider qualitative issues. Moreover, it is noteworthy that quantitative research is missing. We expect that there will be a lot more publications in the coming years. In order to assess and compare those publications with the existing body of knowledge, the present paper provides a fundamental framework based on concepts of RPA.

\section{Summary and Outlook}
\label{ch:summary}

RPA is a novel technology starting to emerge in 2015. By means of an SLR, we provide an overview of the most relevant publications until June 2020. We discovered seven thematic clusters answering fundamental questions such as `What is RPA?', `Which business processes can be automated with RPA?', and `What are the RPA effects?'. Furthermore, we investigate the differences between RPA and related technologies, methods for improving the implementation of RPA projects, and whether AI is used in combination with RPA. Additionally, we provide a review of case studies including the business area, process, and the automation tool.

The paper describes ANCOPUR, a framework for analysing and comparing publications in the RPA area. With the help of criteria, publications can be classified. The framework provides a robust and expandable systematics to categorise and evaluate trends and further developments in the RPA area. Therefore, it will help both scientists and users from industry to assess and compare upcoming RPA publications. 

As discussed, due to the novelty of RPA, the research focus lies on analysing and understanding the RPA technology. The Combination of AI with RPA and the development of Methods for RPA implementation are still in the beginning. Regarding the publication dates of the respective publications, there is a clear trend in this direction visible: nine of ten publications combining RPA and AI and 20 of 22 method papers were published in 2018, 2019, and 2020.

\bibliographystyle{tfcad}
\bibliography{references}

\end{document}